\documentclass{article}

\usepackage{amsmath,amsfonts,amssymb,times,graphicx,natbib,algorithm,algorithmic,hyperref}
\usepackage{booktabs}
\usepackage{subcaption}

\usepackage[accepted]{whi2016}

\def\eg{\textit{e.g}}

\def\ie{\textit{i.e}}

\def\ic{\textit{i.c}}
\def\pc{\textit{p.c}}

\newcommand{\reals}{\mathbb{R}}

\icmltitlerunning{Learning Interpretable Musical Compositional Rules and Traces}

\begin{document}

\twocolumn[
\icmltitle{Learning Interpretable Musical Compositional Rules and Traces}

\icmlauthor{Haizi Yu}{haiziyu7@illinois.edu}
\icmlauthor{Lav R. Varshney}{varshney@illinois.edu}
\icmlauthor{Guy E. Garnett}{garnett@illinois.edu}
\icmlauthor{Ranjitha Kumar}{ranjitha@illinois.edu}
\icmladdress{University of Illinois at Urbana-Champaign, Urbana, IL 61801, USA}

\vskip 0.3in
]

\begin{abstract}
Throughout music history, theorists have identified and documented interpretable rules that capture the decisions of composers. This paper asks, ``Can a machine behave like a music theorist?" It presents MUS-ROVER, a self-learning system for automatically discovering rules from symbolic music. MUS-ROVER performs feature learning via $n$-gram models to extract compositional rules --- statistical patterns over the resulting features. We evaluate MUS-ROVER on Bach's (SATB) chorales, demonstrating that it can recover known rules, as well as identify new, characteristic patterns for further study. We discuss how the extracted rules can be used in both machine and human composition.
\end{abstract}

\section{Introduction}
\label{sec:intro}

For centuries, music theorists have developed concepts and rules to describe the regularity in music compositions. Pedagogues have documented commonly agreed upon compositional rules into textbooks (\eg., Gradus ad Parnassum) to teach composition. With recent advances in artificial intelligence, computer scientists have translated these rules into programs that automatically generate different styles of music \cite{cope1996experiments,biles1994genjam}. However, this paper studies the \emph{reverse} of this pedagogical process, and poses the question: can a machine independently extract from symbolic music data compositional rules that are instructive to both machines and humans?

This paper presents MUS-ROVER, a self-learning system for discovering compositional rules from raw music data (\ie., pitches and their durations). Its rule-learning process is implemented through an iterative loop between a \emph{generative} model --- ``the student" --- that emulates the input's musical style by satisfying a set of learned rules, and a \emph{discriminative} model --- ``the teacher" --- that proposes additional rules to guide the student closer to the target style. The self-learning loop produces a rule book and a set of reading instructions customized for different types of users.

MUS-ROVER is currently designed to extract rules from four-part music for single-line instruments. We represent compositional rules as probability distributions over features abstracted from the raw music data. MUS-ROVER leverages an evolving series of $n$-gram models over these higher-level feature spaces to capture potential rules from both horizontal and vertical dimensions of the texture. 

We train the system on Bach's (SATB) chorales (transposed to C), which have been an attractive corpus for analyzing knowledge of voice leading, counterpoint, and tonality due to their relative uniformity of rhythm \cite{taube1999automatic,rohrmeier2008statistical}. We demonstrate that MUS-ROVER is able to automatically recover compositional rules for these chorales that have been previously identified by music theorists. In addition, we present new, human-interpretable rules discovered by MUS-ROVER that are characteristic of Bach's chorales. Finally, we discuss how the extracted rules can be used in both machine and human composition.

\section{Related Work}
\label{sec:related_work}

Researchers have built expert systems for automatically analyzing and generating music. Many analyzers leverage predefined concepts  (\eg., chord, inversion, functionality) to annotate music parameters in a pedagogical process \cite{taube1999automatic}, or statistically measure a genre's accordance with standard music theory \cite{rohrmeier2008statistical}. Similarly, automatic song writers such as EMI \cite{cope1996experiments} and GenJem \cite{biles1994genjam} rely on explicit, ad-hoc coding of known rules to generate new compositions \cite{merz2014implications}.

In contrast, other systems generate music by learning statistical models such as HMMs and neural networks that capture domain knowledge --- patterns --- from data  \cite{simon2008mysong,mozer1994neural}. Recent advances in deep learning take a step further, enabling knowledge discovery via feature learning directly from raw data \cite{bengio2009learning,bengio2013representation,rajanna2015deep}. However, the learned, high-level features are implicit and non-symbolic with post-hoc interpretations, and often not directly comprehensible or evaluable.

MUS-ROVER both automatically extracts rules from raw data --- without prior encoding any domain knowledge --- and ensures that the rules are interpretable by humans. Interpretable machine learning has studied systems with similar goals in other domains \cite{malioutov2013exact,dash2015learning}.

For readers of varying musical background, the music terminology mentioned throughout this paper can be referenced by any standard textbook on music theory \cite{laitz2012complete}.

\section{The Rule-Learning System}
\label{sec:overview}

We encode the raw representation of a chorale --- pitches and their durations --- symbolically as a four-row matrix, whose entries are MIDI numbers ($C4\mapsto 60, C\sharp 4\mapsto 61$).
The rows represent the horizontal \emph{melodies} in each voice.
The columns represent the vertical \emph{sonorities}, where each column has unit duration equaling the greatest common divisor (gcd) of note durations in the piece.

MUS-ROVER extracts compositional rules, $r = (\phi,p_\phi)$: probability distributions over learned features ($\phi$), from both horizontal and vertical dimensions of the texture.
MUS-ROVER prioritizes vertical rule extractions via a \emph{self-learning loop}, and learns horizontal rules through a series of \emph{evolving $n$-grams}.

\subsection{Self-Learning Loop}

The self-learning loop identifies vertical rules about \emph{sonority} (a \emph{chord} in traditional harmonies) constructions.
Its two main components are ``the student" --- a \emph{generative} model that \emph{applies} rules, and ``the teacher" --- a \emph{discriminative} model that \emph{extracts} rules.
The loop is executed iteratively starting with an empty rule set and an unconstrained student who picks pitches uniformly at random. In each iteration, the teacher compares the student's writing style with Bach's works, and extracts a new rule that augments the current rule set. The augmented rule set is then used to retrain the student, and the updated student is sent to the teacher for the next iteration.

The student in the $k$th iteration is trained by the rule set $R^{\langle k \rangle} = \{r^{(1)}, \ldots, r^{(k)}\}$ via the following optimization problem for the student's probabilistic model $p$:
\begin{align}
\label{eqn:opt-student}
\underset{p \in \Delta^{d}}{\mbox{maximize}}\quad & I(p) \\
\mbox{subject to}\quad 
& p \in \Gamma_1, \ldots, p \in \Gamma_k, \notag
\end{align}
where $p\in\Gamma_i$ requires $p$ to satisfy the $i$th rule $r^{(i)} \in R^{\langle k \rangle}$ with $\Gamma_i$ being the constraint set derived from $r^{(i)}$, and the objective $I: \Delta^{d} \mapsto \reals$ is 
a Tsallis entropy, which achieves a maximum when $p$ is uniform and a minimum when $p$ is deterministic.  Thus the constrained maximization of $I(p)$ disperses probability mass across all the rule-satisfying possibilities and encourages creativity from randomness.  

The teacher in the $k$th iteration solves the following optimization problem for feature $\phi$:
\begin{align}
\label{eqn:opt-teacher}
\mbox{maximize}\quad & s\left(\hat{p}_{\phi|\mathcal{C}_{bach}}, ~p_{\phi}^{\langle k-1 \rangle}\right) \\
\mbox{subject to}\quad & \phi \in \Phi\backslash \left( \Phi^{\langle k-1 \rangle} \cup \left\{\phi_{raw}\right\}\right), \notag
\end{align}
where the objective is a scoring function that selects a feature whose distribution is both regular --- small $\hat{p}_{\phi|\mathcal{C}_{bach}}$ and discriminative --- large $D_{KL}\left(\hat{p}_{\phi|\mathcal{C}_{bach}}, ~p_{\phi}^{\langle k-1 \rangle}\right)$; the constraint signifies that the candidates are the unlearned high-level features. MUS-ROVER highlights the automaticity and interpretability of feature generation. It constructs the universe of all features $\Phi$ via the combinatorial enumeration of selection windows and basis features (descriptors):
\begin{align}
\label{eqn:Phi}
\Phi = \left\{d \circ w ~\middle|~ d \in D, w \in W \right\}.
\end{align}
The descriptor set $D$ is hand-designed but doesn't require domain knowledge, leveraging only basic observations, such as the distance and periodicity, as well as the ordering of the pitches. On the contrary, the window set $W$ is machine enumerated to ensure the exploration capacity. The construction of $\Phi$ guarantees the interpretability for all $\phi \in \Phi$. For instance, people can read out the feature specified by $d_{interv12} \circ w_{\{1,4\}}$ as the piano distance modulo 12 (interval class) between the soprano and bass pitches.

This idea of ``learning by comparison" and the collaborative setting between a generative and discriminative model are similarly presented in statistical models such as noise-contrastive estimation \cite{gutmann2010noise} and generative adversarial networks \cite{goodfellow2014generative}. Both models focus on density estimations to approximate the true data distribution for the purpose of generating similar data; in contrast, our methods do explain the underlying mechanisms that generate the data distribution, such as the compositional rules that produce Bach's styles.

\subsection{Evolving $n$-grams on Feature Spaces}

MUS-ROVER employs a series of $n$-gram models (with \emph{words} being vertical features) to extract horizontal rules that govern the transitions of the sonority features.
All $n$-grams encapsulate copies of self-learning loops to accomplish rule extractions in their contexts. 
Starting with unigram, MUS-ROVER gradually evolves to higher order $n$-grams by initializing an $n$-gram student from the latest ($n$-$1$)-gram student. 
While the unigram model only captures vertical rules such as concepts of intervals and triads, the bigram model searches for rules about sonority progressions such as parallel/contrary motions.

MUS-ROVER's $n$-gram models operate on high-level feature spaces, which is in stark contrast with many other $n$-gram applications in which the words are the raw inputs.
In other words, a higher-order $n$-gram in MUS-ROVER shows how vertical features (high-level abstractions) transition horizontally, as opposed to how a specific chord is followed by other chords (low-level details). Therefore, MUS-ROVER does not suffer from low-level variations in the raw inputs, highlighting a greater generalizability.

\section{Learning Rules from Bach's Chorales}

\subsection{A Rule Book on Bach's chorales}

\begin{figure}[t!]
  \begin{centering}
    \includegraphics[width=0.94\columnwidth]{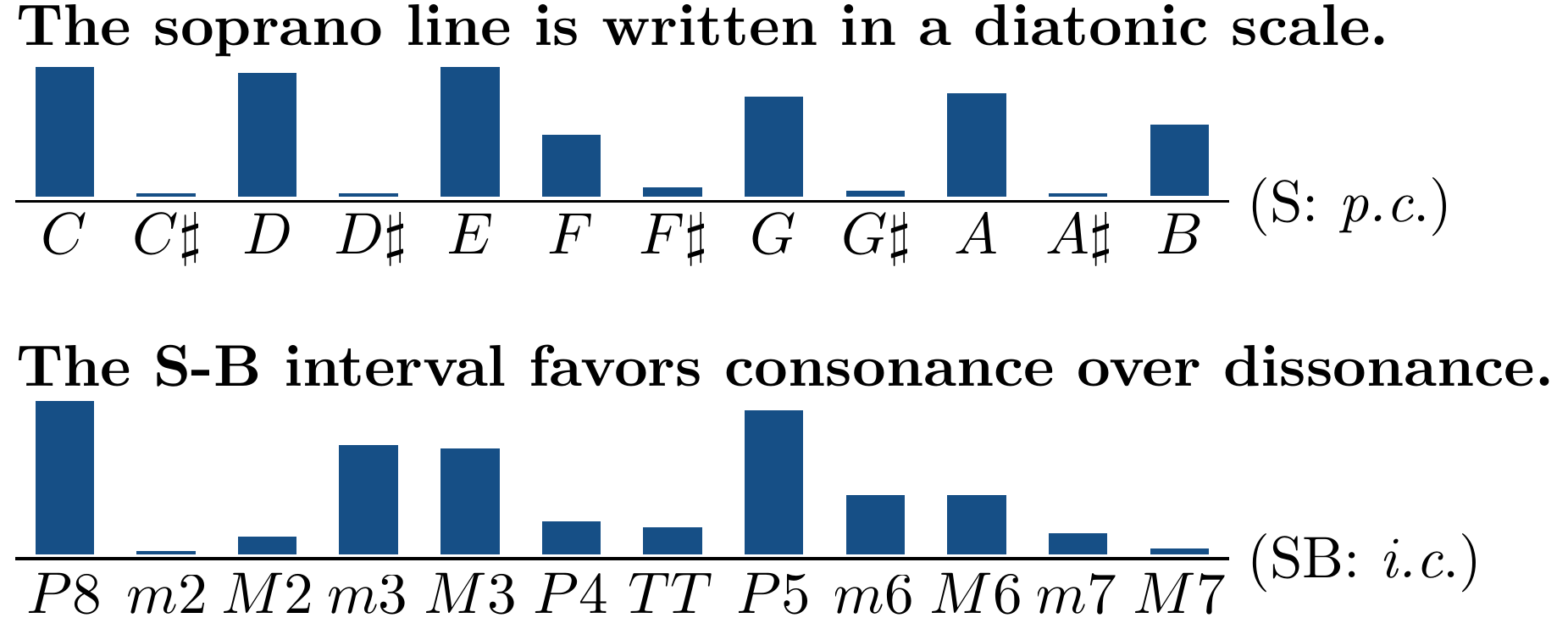}
    \caption{Unigram rule examples from Bach's chorales.}
    \label{fig:unigram-rules}
  \end{centering}
\end{figure}

The rule book contains 63 unigram rules resulting from the feature set $\Phi$, all of which are probability distributions of the associated features. Figure \ref{fig:unigram-rules} illustrates two examples of the unigram rules, whose associated features are $d_{pitch12}\circ w_{\{1\}}$ (top) and $d_{interv12}\circ w_{\{1,4\}}$ (bottom), respectively.
The first example considers the soprano voice, and its descriptor $d_{pitch12}$ is semantically equivalent to \emph{pitch class} (\pc.). It shows the partition of two \pc. sets, which says that the soprano line is built on a diatonic scale. 
The second example considers the soprano and bass, and its descriptor $d_{interv12}$ is semantically equivalent to \emph{interval class} (\ic.). It recovers our definition of intervalic quality: consonance versus dissonance.

\begin{figure}[t!]
  \begin{centering}
    \includegraphics[width=0.94\columnwidth]{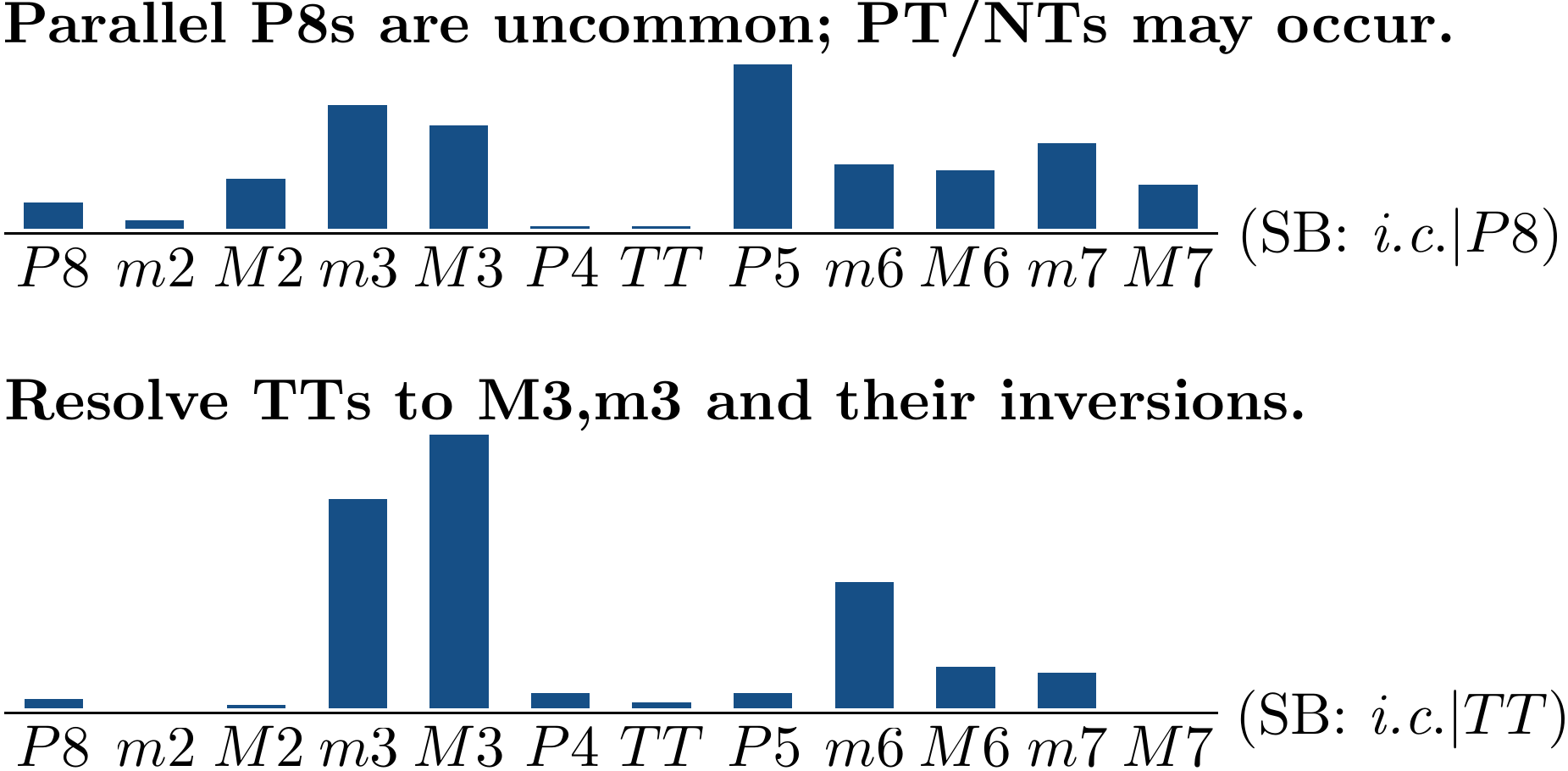}
    \caption{Bigram rule examples from Bach's chorales.}
    \label{fig:bigram-rules}
  \end{centering}
\end{figure}

Given a feature $\phi$, a bigram rule is represented by the feature transition distribution $p_\phi(\cdot|\cdot)$.
Due to the large amount of conditionals for each feature, the book contains many more bigram rules than unigram rules.
Figure \ref{fig:bigram-rules} illustrates two examples of the bigram rules, both of which are associated with feature $d_{interv12}\circ w_{\{1,4\}}$.
Comparing the top bigram rule in Figure \ref{fig:bigram-rules} with the bottom unigram rule in Figure \ref{fig:unigram-rules} shows the re-distribution of the probability mass for feature $d_{pitch12}\circ w_{\{1,4\}}$, the \ic. between soprano and bass (SB: \ic.). The dramatic drop of $P8$ recovers the rule that avoids parallel P8s, while the rises of $m7,M7$ and their inversions suggest the usage of passing/neighbor tones (PT/NTs). The bottom rule in Figure \ref{fig:bigram-rules} illustrates \emph{resolution} --- an important technique used in tonal harmony --- which says tritones (TTs) are most often resolved to $m3,M3$ and their inversions. 
Interestingly, the fifth peak ($m7$) in the pmf of this rule reveals an observation that doesn't fall into the category of resolution. This transition, $TT\rightarrow m7$, is similar to the notion of escape tone (ET), which suspends the tension instead of directly resolving it. For instance, $(F4,B2) \rightarrow (F4,G2)$, which will eventually resolve to $(E4,C3)$.
All of these rules are automatically identified during rule-learning. 

\subsection{Customized Rule-Learning Traces}

Despite its readability for every single rule, the rule book is in general hard to read as a whole due to its length and lack of organization. 
MUS-ROVER's self-learning loop solves both challenges by offering customized \emph{rule-learning traces} --- ordered rule sequences --- resulting from its iterative extraction. Therefore, MUS-ROVER not only outputs a comprehensive rule book, but more crucially, suggests ways to read it, tailored to different types of students.

We propose two criteria, \emph{efficiency} and \emph{memorability}, to assess a rule-learning trace from the unigram model. The efficiency measures the speed in approaching Bach's style; the memorability measures the complexity in memorizing the rules. A good trace is both efficient in imitation and easy to memorize.

To formalize these two notions, we first define a rule-learning trace $\mathcal{T}^{\langle k \rangle}$ as the ordered list of the rule set $R^{\langle k \rangle}$,
and quantify the gap against Bach by the KL divergence in the raw feature space:
$gap^{\langle k \rangle} = D(~\hat{p}_{\phi_{raw}|\mathcal{C}_{bach}} ~\|~ p_{\phi_{raw}}^{\langle k \rangle}~)$.
The \emph{efficiency} of $\mathcal{T}^{\langle k\rangle}$ with efficiency level $\epsilon$ is defined as the minimum number of iterations that is needed to achieve a student that is $\epsilon$-close to Bach if possible:
\[
\mathsf{E}_\epsilon \left(\mathcal{T}^{\langle k \rangle}\right) = 
\begin{cases}
\min \left\{n ~\middle|~ gap^{\langle n \rangle}<\epsilon\right\}, & gap^{\langle k \rangle} < \epsilon; \\
\infty, & gap^{\langle k \rangle} \geq \epsilon.
\end{cases}
\]
The \emph{memorability} of $\mathcal{T}^{\langle k\rangle}$ is defined as the average entropy \cite{pape2015complexity} of the feature distributions from the first few efficient rules:
\[
\mathsf{M}_\epsilon \left(\mathcal{T}^{\langle k \rangle}\right) = \frac{1}{N} \sum_{k=1}^{N} H\left(\hat{p}_{\phi^{(k)}|\mathcal{C}_{bach}}\right),
\]
where $N = \min\left\{k,\mathsf{E}_\epsilon \left(\mathcal{T}^{\langle k \rangle}\right)\right\}$. 
There is a tradeoff between efficiency and memorability. At one extreme, it is most efficient to just memorize $p_{\phi_{raw}}$, which takes only one step to achieve a zero gap, but is too complicated to memorize or learn. At the other extreme, it is easiest to just memorize $p_\phi$ for ordering related features, which are (nearly) deterministic but less useful, since memorizing the orderings takes you acoustically nowhere closer to Bach.
The $\alpha$ parameter in the scoring function of \eqref{eqn:opt-teacher} is specially designed to balance the tradeoff, with a smaller $\alpha$ for more memorability and a larger $\alpha$ for more efficiency (Table \ref{tab:efficiency-and-memorability}).

\begin{table}[t!]
  \centering
  \begin{tabular}{crrr}
    \multicolumn{4}{c}{\includegraphics[width=0.94\columnwidth]{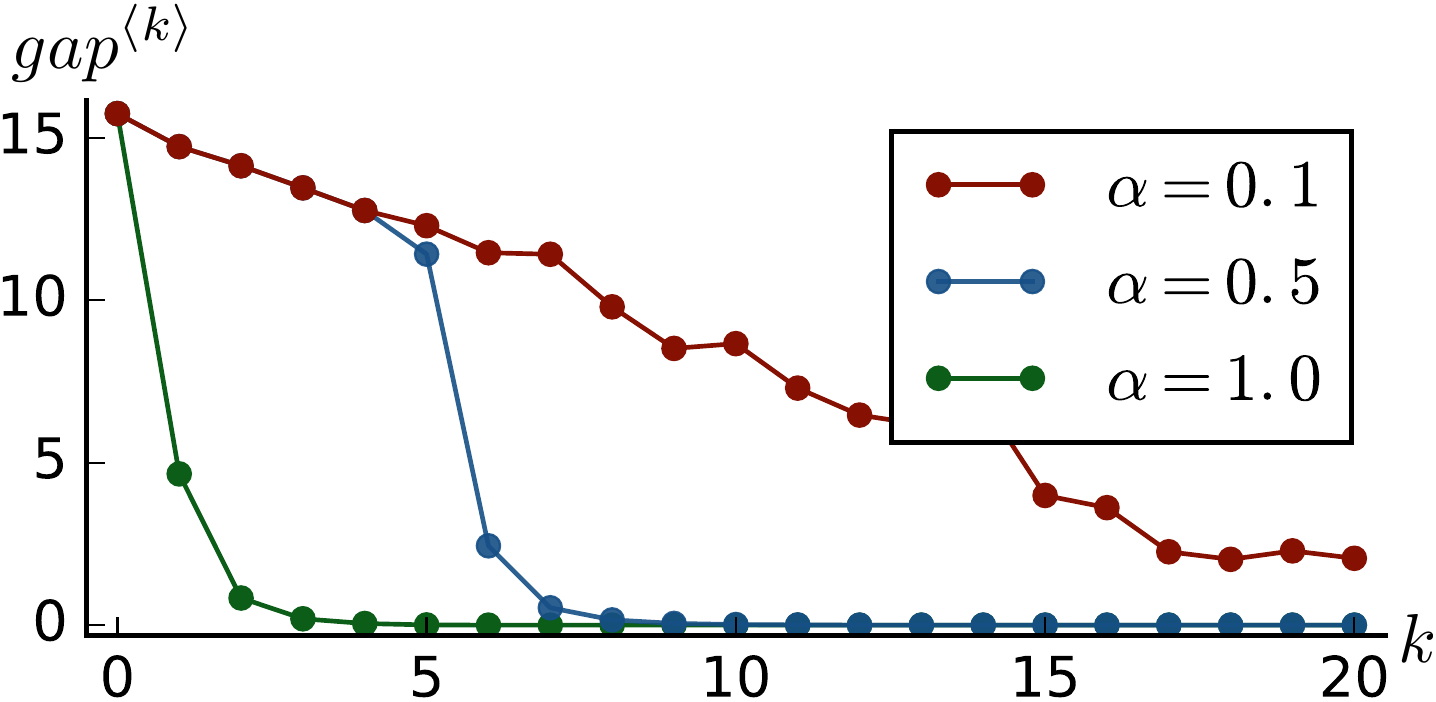}} \\
    & & \multicolumn{2}{c}{\small{\textbf{Unigram Rule-Learning Traces}}} \\
    \cmidrule(r){3-4}
    & {\small$\alpha = 0.1$} & {\small$\alpha = 0.5$} & {\small$\alpha = 1.0$} \\
    \midrule
    1 & (1,4), order & (1,4), order & (1,2,3), pitch \\
    2 & (1,3), order & (1,3), order & (2,3,4), pitch \\
    3 & (2,4), order & (2,4), order & (1,2,3,4), pitch12 \\
    4 & (1,2), order & (1,2), order & (1,3,4), pitch \\
    5 & (2,3), order & (2,3,4), order & (1,2,4), pitch \\
    6 & (3,4), order & (1,3,4), pitch & (1,2,3,4), interv \\
    $\cdots$ & $\cdots$ & $\cdots$ & $\cdots$ \\
    \midrule
    $\mathsf{E}_\epsilon $ & $\infty$ & $12$ & $6$  \\
    $\mathsf{M}_\epsilon $ & $2.21$ & $4.97$ & $8.63$ \\
  \end{tabular}
  \caption{Three unigram rule-learning traces ($\mathcal{T}^{\langle 20 \rangle}$) with $\alpha = 0.1,0.5,1.0$. The top figure shows the footprints that mark the diminishing gaps. The bottom table records the first six rules, and shows the tradeoff between efficiency and memorability ($\epsilon = 0.005$). The trace with $\alpha = 1.0$ shows the most efficiency, but the least memorability.}~\label{tab:efficiency-and-memorability}
\end{table}

\begin{figure}[t!]
    \centering
    \begin{subfigure}[b]{0.49\columnwidth}
        \includegraphics[width=\textwidth]{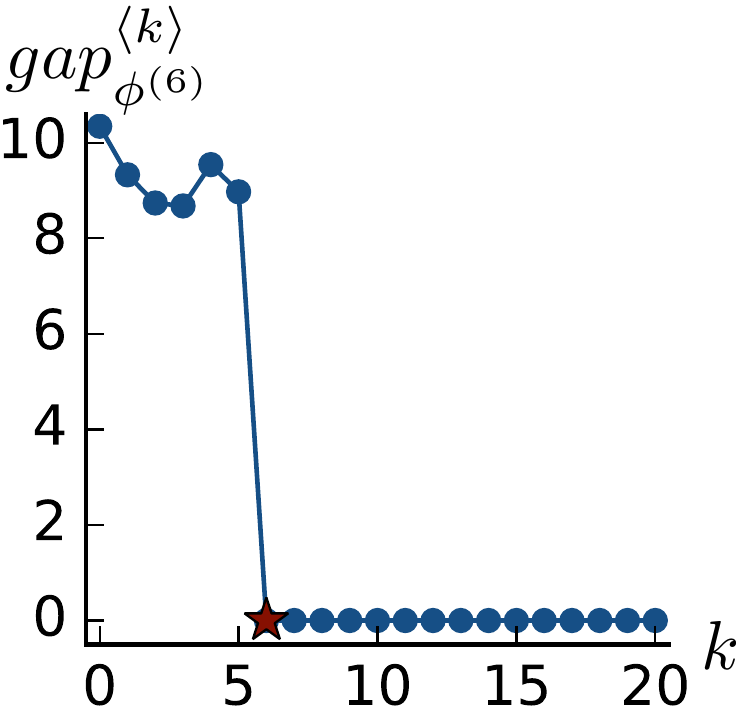}
        \caption{$\phi^{(6)} = d_{pitch}\circ w_{\{1,3,4\}}$}
        \label{fig:feat6_134_pitch}
    \end{subfigure}
    \begin{subfigure}[b]{0.49\columnwidth}
        \includegraphics[width=\textwidth]{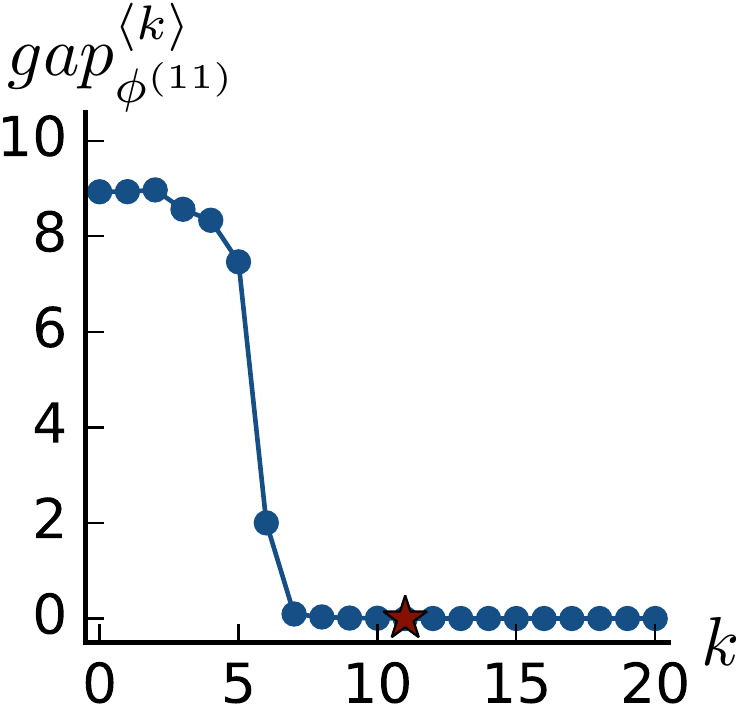}
        \caption{$\phi^{(11)} = d_{interv}\circ w_{\{1,2,3,4\}}$}
        \label{fig:feat11_1234_interv}
    \end{subfigure}
    \caption{Rule entanglement: two sets of footprints that mark the diminishing gaps, both of which are from the rule-learning trace with $\alpha=0.5$. The location of the star shows whether the associated rule is entangled (right) or not (left).}\label{fig:entanglement}
\end{figure}

To study the rule entangling problem, we generalize the notion of $gap$ from the raw feature to all high-level features: 
\[
gap_\phi^{\langle k \rangle} = D(~\hat{p}_{\phi|\mathcal{C}_{bach}} ~\|~ p_\phi^{\langle k \rangle}~), \quad \forall \phi \in \Phi.
\]
Plotting the footprints of the diminishing gaps for a given feature reveals the (possible) implication of its associated rule from other rules. For instance, Figure \ref{fig:entanglement} shows two sets of footprints for $\phi^{(6)}$ and $\phi^{(11)}$. 
By starring the iteration when the rule of interest is actually learned, we see that $r^{(6)}$ cannot be implied from the previous rules, since learning this rule dramatically closes the gap; on the contrary, $r^{(11)}$ can be implied from the starting seven or eight rules.

Given a rule-learning trace in the bigram setting, the analysis on efficiency and memorability, as well as feature entanglement, remains the same. However, every trace from the bigram model is generated as a continuation of unigram learning: the bigram student is initialized from the latest unigram student. This implies the bigram rule set is initialized from the unigram rule set, rather than an empty set. MUS-ROVER uses the extracted bigram rules to overwrite their unigram counterparts --- rules with the same features --- highlighting the differences between the two language models. The comparison between a bigram rule and its unigram counterpart is key in recovering rules that are otherwise unnoticeable from the bigram rule alone, such as ``Parallel P8s are avoided!" Thus, MUS-ROVER emphasizes the necessity of tracking a series of evolving $n$-grams, as opposed to learning from the highest possible order only.

\section{Discussion and Future Work}
\label{sec:discussions}


MUS-ROVER takes a first step in automatic knowledge discovery in music, and opens many directions for future work. Its outputs --- the rule book and the learning traces --- serve as static and dynamic signatures of an input style. We plan to extend MUS-ROVER beyond chorales, so we can analyze similarities and differences of various genres through these signatures, opening opportunities for style mixing. Moreover, while this paper depicts MUS-ROVER as a fully-automated system, we could have a human student become the generative component, interacting with ``the teacher" to get iterative feedback on his/her compositions.

A more detailed version of this work will appear elsewhere \cite{yuvgk2016}.

\section*{Acknowledgements} 
 
We thank Professor Heinrich Taube, President of Illiac Software, Inc., for providing Harmonia's MusicXML corpus of Bach's chorales.\footnote{\url{http://www.illiacsoftware.com/harmonia}}

\bibliography{mus-rover-whi-fin}
\bibliographystyle{icml2016}

\end{document}